\documentclass[conference]{IEEEtran}
\IEEEoverridecommandlockouts
\usepackage{cite}
\usepackage{amsmath,amssymb,amsfonts}
\usepackage{algorithmic}
\usepackage{graphicx}
\usepackage{textcomp}
\usepackage{float}
\usepackage{subfig}
\usepackage{hyperref}
\usepackage{float}
\usepackage{multirow}
\usepackage[table,xcdraw]{xcolor}
\def\BibTeX{{\rm B\kern-.05em{\sc i\kern-.025em b}\kern-.08em
    T\kern-.1667em\lower.7ex\hbox{E}\kern-.125emX}}
\begin{document}

\title{Image Segmentation with Topological Priors\\}

\author{\IEEEauthorblockN{Shakir Showkat Sofi}
\IEEEauthorblockA{\textit{CDISE,} \\
\textit{Skolkovo Institute of Science and Technology}\\
Moscow, Russia \\
Shakir.Sofi@skoltech.ru}
\and
\IEEEauthorblockN{Nadezhda Alsahanova}
\IEEEauthorblockA{\textit{CDISE,} \\
\textit{Skolkovo Institute of Science and Technology}\\
Moscow, Russia \\
nadezhda.alsahanova@skoltech.ru}
}

\maketitle

\begin{abstract}
Solving segmentation tasks with topological priors proved to make fewer errors in fine-scale structures. In this work, we use topological priors both before and during the deep neural network training procedure. We compared the results of the two approaches with simple segmentation on various accuracy metrics and the Betti number error, which is directly related to topological correctness, and discovered that incorporating topological information into the classical UNet model performed significantly better. We conducted experiments on the ISBI EM segmentation dataset.
\end{abstract}


\section{Introduction}
The important task in computer vision is to know the location and shape of objects in the image or in general, which pixel belongs to which object. This task is accomplished through image segmentation, which involves assigning labels to all input image pixels. The use of an end-to-end trained deep network to segment images aids in achieving acceptable per-pixel accuracy. However, in applications involving fine-scale structures, such as thin connections in neuron membranes and vessels, satisfactory accuracy is insufficient because it could lead to catastrophic mistakes. A segmentation error in the thin cell membrane, for example, could result in the union of two distinct cells. Thus,  segmentation algorithms can still make mistakes on fine-scale structures. To solve this problem, we incorporate topological prior knowledge into the segmentation model.

In this work, we investigate two approaches to introducing topological priors. The first is to add topological loss to cross-entropy loss, which is commonly used in segmentation tasks. Topological loss measures a difference between persistence diagrams for true and predicted masks. The second method is to use topological image processing prior to training the neural network. Both strategies improve segmentation performance without sacrificing pixel-wise accuracy.

In section \ref{Related work} we present a short review of existing solutions for the implementation of topological priors in segmentation tasks. Then in section \ref{Theory} we present a short theoretical basis of persistence homology in\ref{Persistent homology}, topological loss and its differentiability is discussed in  \ref{Topological Loss} and the overview of topological input image processing is presented in  \ref{Topological image processing}. Finally, we present details of the training process and result of experiments in section \ref{Experiments}.

\section{Related Work}
\label{Related work}
Topologically aware networks have already shown significant improvement in results, especially for segmentation and classification problems.  One such earlier attempt was in \cite{mosin}, which uses topological awareness in the loss function based on the response of selected filters from a pre-trained VGG19 network.  It was successful in capturing some topological features, such as the connectedness of small components. It constructs the topologically aware losses, but it is a bit difficult to generalize for complex settings. The interpretation and relevance of these captured features were even more difficult. A similar scheme was proposed by researchers in \cite{okty},  where the output of the second network was used to define a loss function for identifying global structural features, to enforce anatomical constraints.  Some researchers, like in \cite{Chen}, uses topological regularizers for classification problems with considerations of stability of connected components within imposed topological constraints on the shape of classification boundary. Different ideas have been proposed so far for capturing fine details, some based on deconvolution and upsampling,  some using Persistent Homology (PH), and some have used topological processing of inputs for simple geometries for unsupervised tasks like in \cite{tiip} before applying the {\em Chan-Vese, ISODATA, Edge-detections etc} for segmenting the processed image.  For further see \cite{assaf}; \cite{clough1};  \cite{clough2}; \cite{ying}, etc. Most of the methods are problem-specific. Our method is closest to one proposed in \cite{xiao1}, which also poses topological priors in the training phase. Apart from that, our scheme looks at different possible stages of feeding topological priors with their effectiveness and efficiency for supervised learning task.

\section{Theory}
\label{Theory}
We know from recent years that the volume of data is increasing exponentially, but it is also complex, noisy, multidimensional, and sometimes incomplete, necessitating the development of efficient and robust data analysis methods.  The existing methods, which are based on statistics, machine learning, uncertainty quantification, and so on, work well, but when it comes to making sense of vast, multidimensional, and noisy data, the analysis process becomes more challenging. However, developments in modern mathematics have yielded a wealth of insights into the study and application of data in an entirely new set of directions. Topological Data Analysis (TDA) is one such technique that combines computational geometry, algebraic topology, data analysis, statistics, etc. TDA is more involved in the geometric side of data, like computing topological features of data. Furthermore, the prime focus is on Persistent Homology (PH). We provide a brief overview of PH here, but the reader is directed to  \cite{Edelph}; \cite{Otterroadmap} for a more in-depth discussion.

\subsection{Persistent homology and Persistent diagrams}
\label{Persistent homology}
A persistent homology is an important tool of algebraic topological used in TDA to analyze qualitative data features across multiple scales. PH is robust, dimension and coordinate independent, and provides compact summaries of the underlying data. To compute PH, we need two things: a simplicial complex $K$ and the filtration $\mathcal{F}$ defined on $K$.   A simplicial complex is a generalisation of a graph that includes 0-simplices (nodes), 1-simplices (edges), 2-simplices (triangles), and so on up to $k$-simplices, where $k \in \mathbb{N}$ denotes the dimensionality of the complex. It is formally a space that is built from a union of these nodes, edges, triangles, and/or higher-dimensional polytopes \cite{Otterroadmap}. A simplicial complex $K$ is closed under inclusion, which means that if $\sigma^{\prime} \subseteq \sigma \in K$ then $\sigma^{\prime} \in K.$  A nested sequence of $K_{0} \subseteq K_{1}  \subseteq \ldots \subseteq K_{N}=K$ of subcomplexes of $K$ is then defined as the filtration $\mathcal{F}$ on $K$, please refer to \cite{tiip} for details.\par

Due to the grid-like structure of images, it became easy to use TDA techniques. Considering an image $I$ of size $M \times N$, the simplicial complex can be obtained from pixel configurations of this image. Technically, the simplicial complex (1-dim simplicial) is defined by connecting each pixel to 8 neighbouring pixels. After that, by filling the triangular shapes of this complex, a 2-dimensional simplicial complex can be obtained. Assuming that images have a uniform grid structure, we may prefer the {\em cubical complex\footnote{Space formed by the union of vertex, edges, squares, cubes, and so on.}}\par
PH measures lifetimes of topological features through the birth and death of holes across the filtration \cite{tiip}. For example,  if any hole persists for long consecutive values of the varying parameter(e.g., diameter $\tau$ of data-point), then that depicts an important object in an image, and small persistent components are noise. For each complex $K_i$, we compute its topology using Homology group $\mathcal{H}_{n}$\footnote{We don't need all details of computations of Homology groups, as it is already implemented in \texttt{Gudhi Library}}, the rank of which is $k^{th}$ Betti number ($\beta_k$), represents a $k$-dimensional hole. $\beta_0$ is the number of connected components, $\beta_1$ is the number of loops, $\beta_2$ is the number of hallow-cavities, and so on \cite{clough1}. In the case of image data sets, however, we only go up to 1-dimensional holes. There is a pretty good method for visualising and interpreting births and deaths of these holes using {\em Persistent Diagram (PD).} PD is simply a multiset that contains a point $(b, d)$ for every hole that was born at data point diameter $\tau=b$ and died at data point diameter $\tau=d$. It is a plot between the birth and death of these holes for different values of data point diameter $\tau$.

\subsection{Topological Loss and Differentiability}
\label{Topological Loss}
It has been demonstrated that training a deep neural network, typically a TopoNet, according to the methodology proposed in \cite{xiao1} can achieve both per-pixel accuracy and topological correctness. Lets likewise define $\mathbf{f},$ the likelihood map predicted by the network, and $\mathbf{g}$ the ground truth. The overall loss function can be expressed in mathematical terms as the weighted sum of the cross-entropy loss and the topological loss:
\begin{equation}\label{eq0}
    \mathbf{L(f,g) = L_{BCE}(f,g) + \lambda \cdot L_{topo}(f,g)}
\end{equation}  
A binary segmentation task is assumed. Consequently, there is a single likelihood function  $\mathbf{f}$, whose value ranges from 0 to 1.

We use information from persistent diagrams of $\mathbf{f}$ and $\mathbf{g}$, $\mathbf{D(f), D(g)}$, to formalise topological loss, which measures the topological similarity between the likelihood $\mathbf{f}$ and the ground truth $\mathbf{g}$. Mathematically we can write  topological loss as:
\begin{equation} \label{eq1}
\begin{split}
\mathbf{L_{topo}(f,g)} = \sum_{(p,p') \in \mathbf{D(f),D(g)}} &\Big[(\texttt{birth}(p) - \texttt{birth}(p'))^2 + \\
            &+ (\texttt{death}(p) - \texttt{death}(p'))^2 \Big]
\end{split}
\end{equation}
Where $p$ and $p'$ are the points from persistence diagrams $\mathbf{D(f), D(g)}$ sorted by their 'lifetimes':
$$\texttt{death($p$) - birth($p$)}$$
So, in equation  (\ref{eq1}), this loss measures the difference between persistent diagrams. It is dependent on the critical thresholds at which topological changes take place, such as the birth and death times of the various dots depicted in the diagram $\mathbf{D(f)}$ \cite{ 
xiao1}.  And if $\mathbf{f}$ is differentiable, these vital thresholds are critical points where the derivative of $\mathbf{f}$ equals zero. As a consequence of this, it is possible to represent $\mathbf{f}$ as a piecewise-linear function that has a gradient equal to zero on critical points. So, $\mathbf{f}$ is differentiable.
Differentiability of loss function in equation (\ref{eq1}) follows from differentiability of  $\mathbf{f}$. So, gradient of loss function $\mathbf{\nabla_w L_{topo}(f,g) }$ can be written as follows:
\begin{equation} \label{eq2}
\begin{split}
\sum_{(p,p') \in \mathbf{D(f),D(g)}} \Big[ &2(\texttt{birth}(p) - \texttt{birth}(p')) \frac{\partial \mathbf{f}(c_b(p))}{\partial w} + \\
            &+2 (\texttt{death}(p) - \texttt{death}(p'))\frac{\partial \mathbf{f}(c_d(p))}{\partial w}\Big]    
\end{split}
\end{equation}
Where, for each dot $p \in \mathbf{D(f)}$ , we designate the birth and death critical points of the related topological structure by the notation $c_b(p)$ and $c_d(p)$, respectively \cite{xiao1}. Therefore, the gradient can be straightforwardly computed based on the chain rule.

\subsection{Topological input image processing}
\label{Topological image processing}
As previously stated, topological priors can be fed into the segmentation process at many stages, such as the input stage, the training stage, or others.  In the above section, we have seen that when we are posing topological priors during the training phase,  the process becomes sluggish and computationally expensive. To address this issue, we use a new technique (in UNet architecture) that processes the image using topological information before applying the actual segmentation algorithm, removing irrelevant objects and noise and ensuring that the modified image has well-defined topological characteristics. The primary benefit of utilising this method is that it enables us to perform training with a simple cross-entropy loss function, and it also enables us to perform pre-processing on images whenever we like prior to solving the actual segmentation problem. Therefore, it should come as no surprise that the method is effective from a computational standpoint. In addition to this, it is simple to apply to unsupervised segmentation problems such as those found in \cite{tiip}, which would not have been achievable if we had used priors during the training phase. The purpose of this work is to incorporate topological image processing into supervised learning.

\begin{figure}[H]
\vskip 0.05in
\begin{center}
\centerline{\includegraphics[width=\columnwidth]{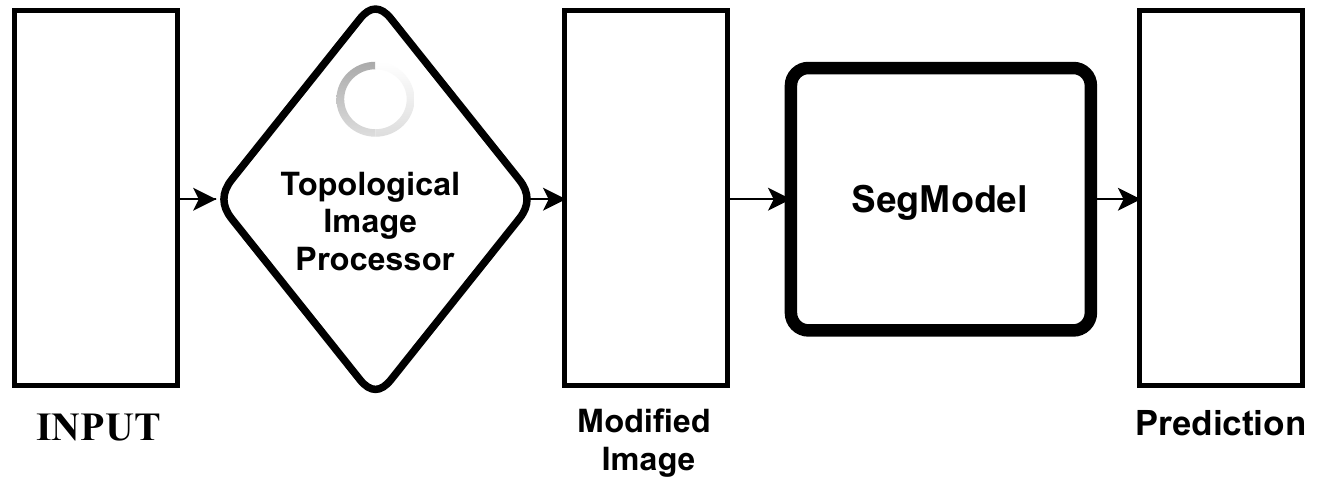}}
\caption{Topological input image processing for segmentation}
\label{tiip}
\end{center}
\vskip -0.05in
\end{figure}
The first type of processing of the input image is {\em image smoothening} since this method has previously shown the advantages with persistent homology \cite{assaf2}. Smoothening can be used to reduce noise, remove pixel-wise focus, to improves quality.  It can be performed by spatial or frequency filters, here in our case,  we use a \texttt{uniform filter} from the standard \texttt{scipy} library, which moves the filter mask of size, say $k \times k$, and replaces each pixel value by the average of its neighbors, including itself. We know the real world images; some images may or may not have borders or other irrelevant objects. So,  in general, it becomes difficult to guarantee that most persistent components are important features in the image. The idea of {\em border modification} was proposed in \cite{tiip}, in which we suppose that the objects of interest in many real-world images do not link with borders, or, in a weaker sense, that most images have objects of interest near the centre. The border processing, in technical terms, constructs an image $I_{b}$ from an original image by ensuring that each pixel within a distance $d$ of the $I_{b}$ border reaches the minimum value, while the rest values remain fixed. From elder rule\footnote{The elder rule asserts that when constructing a persistence diagram, if two components or holes are merged, the youngest component or hole dies. This rule is also known as the {\em old survive rule}\cite{tiip}}, this assures that every object that connects to this border will be born through it. In the PD of $I_{b}$, all irrelevant portions correspond to a single point with infinite persistence, while features have finite persistence.\par
{\em Topological input image processing} is simply processing the images using the geometric information we have. Here, we will use these as processing steps, but there can be plenty of such modifiers.\par
After the first step,  which is to determine the number of components shown by the input image, using lifetime distribution.  Border modification may restrict to finite lifetimes. We may also use any outlier detector for threshold selections. According to the method described in \cite{Rieck}, pertinent peaks can be retrieved from the persistence diagram if the diagram includes a band of a certain width that does not include any points. In this work, we apply this method. In a more technical sense, it is the largest empty region parallel to the diagonal. We are able to draw it into the persistence diagram by simply iterating through lifetimes in decreasing order in order to track the difference between consecutive lifetimes \cite{tiip}.  Finally, that threshold is selected, which gives the largest difference between two lifetimes. After selecting the threshold, we increase the contrast between the objects in the image with a lifetime above that threshold and background. For, marking objects in image $I$, we use the below algorithm, for details please see \cite{Rieck};\cite{tiip}
\begin{figure}[H]
    \centering
    \includegraphics[scale=0.5,width=0.99\linewidth]{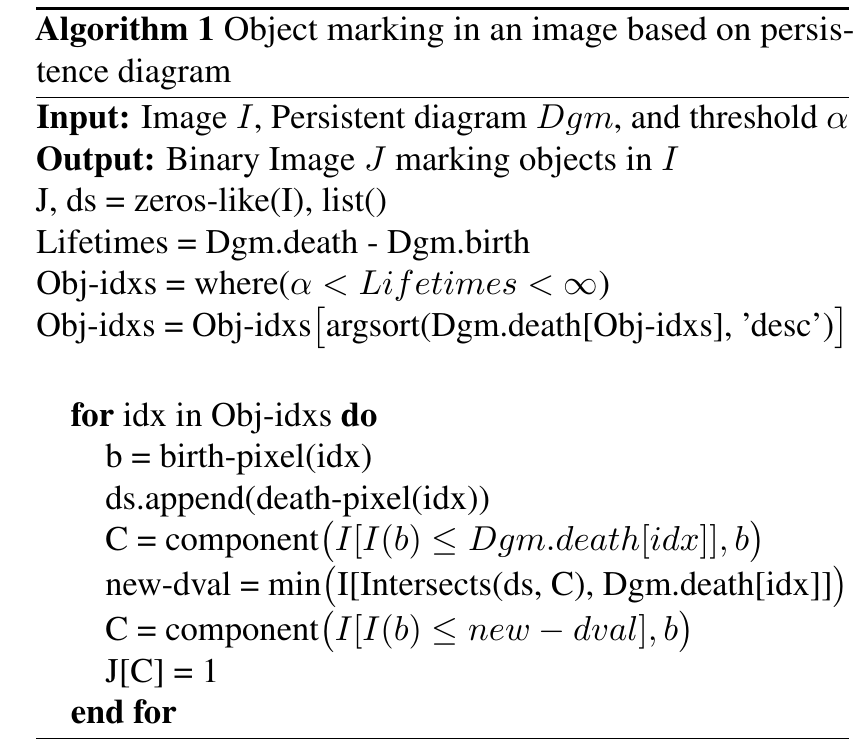}
  \label{falgo}
\end{figure} 

Conceptually, after obtaining the lifetime of components directly from the persistent diagram, we identifying the diagram point having a significant finite lifetime using threshold, then sorting these identified points by decreasing death times. For all sorted point, we first identify the image pixel that corresponds to this diagram point and store the image pixel corresponding to the death time of this diagram point, then in 'C' put all pixels connected to 'b' before its death, ensuring that 'C' doesn't have overlap with previous components, finally mark the component for this diagram point in output, and iterate for all points like-wise. \par
In the end, we use multi-variate interpolation to fill in the background pixels in order to get an image with a smooth transition between the background and the objects in the image,  likewise proposed in \cite{tiip}, but here we have supervised mechanism for training. 
\section{Experiments}
\label{Experiments}
\subsection{Dataset}

The first challenge on 2D segmentation of neuronal processes in EM images begun in May 2012. They have provided training data consisting of 30 image stacks. These images represent a set of consecutive slices within one 3D volume, basically which contains a set of 30 sequential sections from a {\em serial section Transmission Electron Microscopy (ssTEM)} data set of the Drosophila first instar larva ventral nerve cord (VNC). They have provided the ground-truth images as well. Below there is the sample of the image and corresponding mask. It can be easily seen the complexity of the geometry of this dataset. Each training image is of size $512 \times 512$ in grayscale, with corresponding ground truth.  The first pre-processing was to divide the dataset into training and evaluation sets into an 80:20 ratio. Then, we saved the final dataset for which we did all our analysis. Choosing this dataset was a great idea because of its complex geometry, so it was challenging to use the topological priors in this data.
\begin{figure}[H]
    \centering
    \subfloat{\includegraphics[scale=0.5,width=0.48\linewidth]{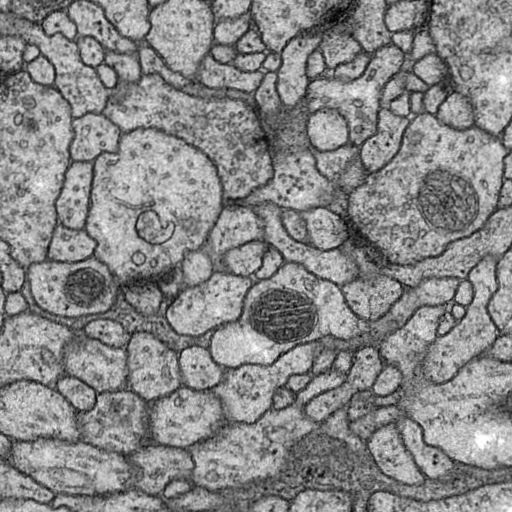} }%
    \subfloat{\includegraphics[scale=0.5,width=0.48\linewidth]{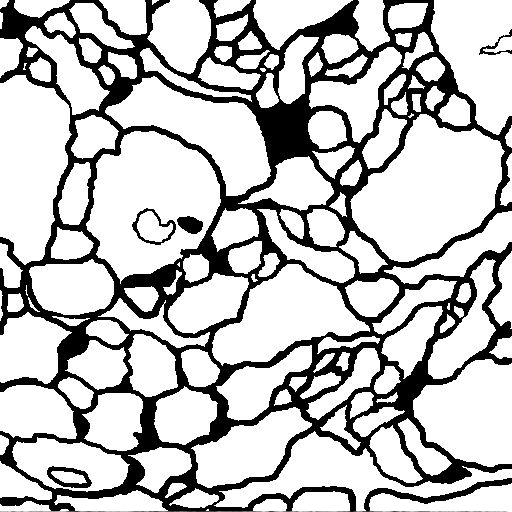}}%
  \caption{Image and corresponding Mask}
\end{figure}
\subsection{Evaluation metrics}
As the segmentation task is already well developed in deep learning literature, it gave us much room for using already proposed metrics as well;  we use a variety of evaluating metrics in our research expository, some of them follows from  \cite{Heip}, where they were used for a similar problem. In addition, average accuracy, correctness, quality, dice, many more,   we have used the topological relevant metric,  i.e., \textbf{Betti number error}, which is directly comparing the topology of prediction and ground-truth. Below we can see the basic definitions, \\
\begin{itemize}
\item \underline{Accuracy:} This is the most used metric for binary segmentation problems, which simply tells us the percentage of correctly classified pixels.

    $$ Accuracy  =  \frac{T P + T N}{T P+ T N+F P+F N}, \in [0; 1]$$

\item \underline{Sørensen-Dice Coefficient:} This metric is the statistical assessment for similarity of two given samples.
$$ Dice  =  \frac{2 TP}{2 T P+F P+F N}, \in [0; 1]$$

\item \underline{ Completeness/lnclusion score/Recall:} Assessing how well the predicted encompasses the ground-truth.
$$Completeness =\frac{T P}{T P+F N}, \in [0;1]$$
 \item \underline{ Correctness/Precision: } The correctness represents the percentage of correctly extracted pixels
$$Correctness =\frac{T P}{T P+F P}, \in [0;1]$$
\item \underline{ Quality :} Quality combines  completeness and correctness into a single measure, so it is more general and is defined as:
$$Quality = \frac{T P}{T P+F P+F N}, \in [0;1]$$

All the above metrics have optimal value 1

\item \underline{Betti number-Error:} This metric is more important in our case as our main focus is about topological segmentation, the betti-number error is more topology-relevant. In this, we randomly sample the small patches of ground-truth and prediction, then compute the average absolute difference of their betti-numbers $\beta_k$  (give us a measure of the number of handles, closed loops etc in that patch), this reveals how much the geometry of prediction is different from ground-truth. The smaller the betti-number error, the better is segmentation, ideal value is zero, which means the prediction map has exactly the same topology as that of ground-truth, sometimes called {\em topologically correct.}
\end{itemize}

\subsection{Details of training procedure}

\begin{figure}[H]
    \centering
    \includegraphics[width=0.97\linewidth]{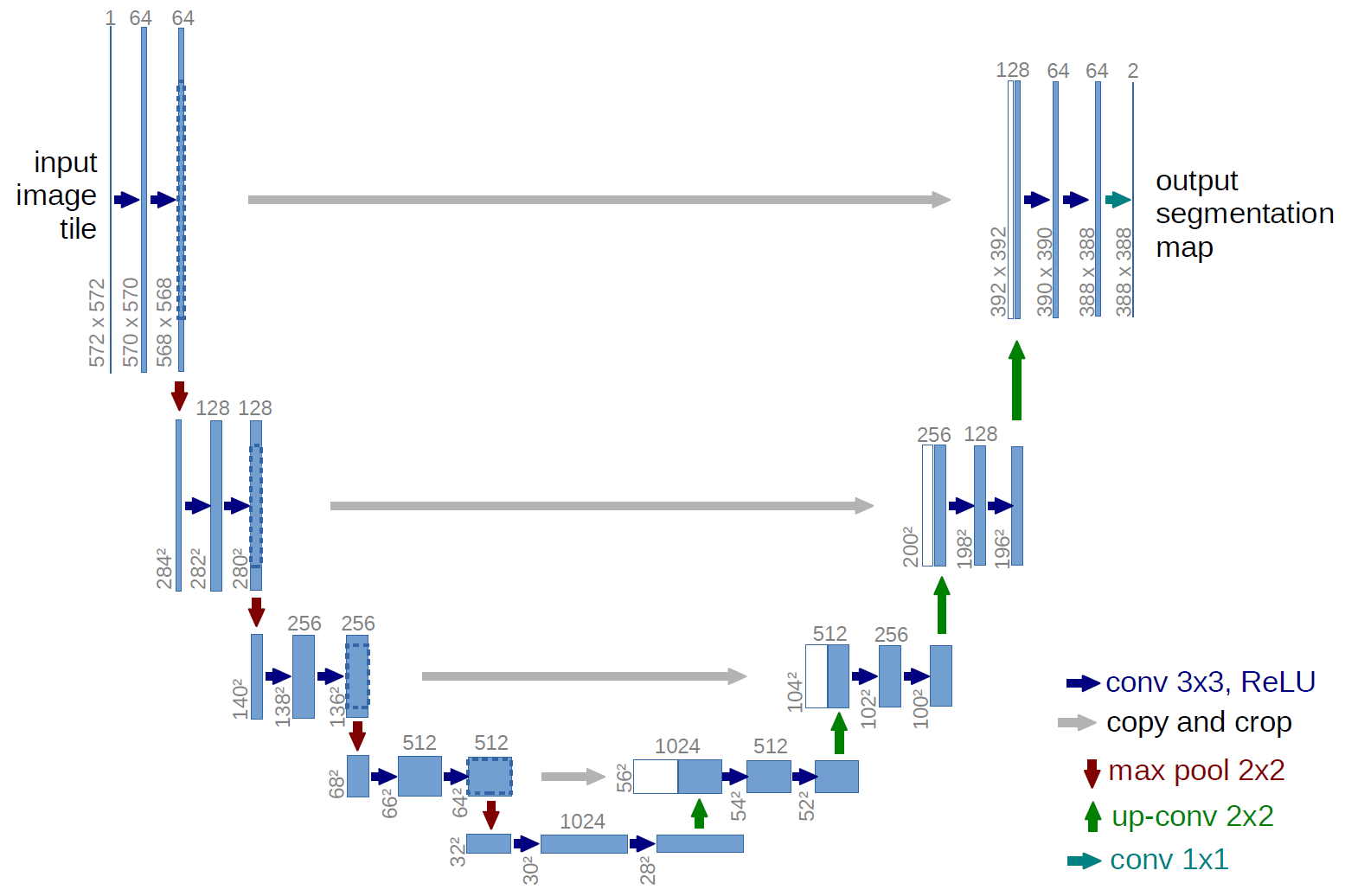}
  \caption{Architecture of U-Net}
  \label{f:1}
\end{figure}

Many deep neural networks are used for the segmentation problem. However, for the task of biomedical segmentation, the best results are shown by U-Net (Fig.\ref{f:1}).
We trained U-Net on augmented small patches from images ($64 \times 64$) as computing topological priors on small patches is faster. Augmentation was used because we have a small dataset, and random choice of patches and their random flipping helps to enlarge the dataset for training. Because of computational complexity, we also take batch size equal to 1. On such a small batch size U-Net with and without topological priors gave better results. 

For the realization of topological loss, we used LevelSetLayer2D function from TopologyLayer \footnote{\href{https://github.com/bruel-gabrielsson/TopologyLayer}{TopologyLayer} is a library where by using of PyTorch library and C++ the differentiable persistence diagram was implemented.}, which calculate persistence diagrams. We added topological loss to CrossEntropy loss \ref{eq0} with $\lambda = \frac1{12000}$ to make this losses have the same order.  Also, for visualization of diagrams, we used Ripser library.

\begin{figure}[H]
    \centering
    \subfloat{\includegraphics[scale=0.5,width=0.48\linewidth]{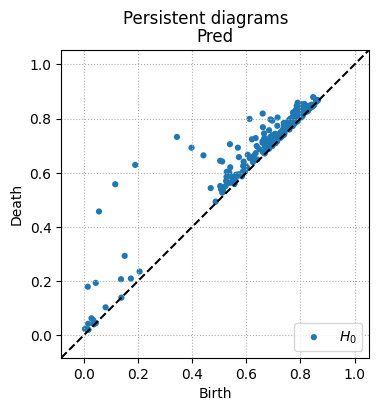} }%
    \subfloat{\includegraphics[scale=0.5,width=0.48\linewidth]{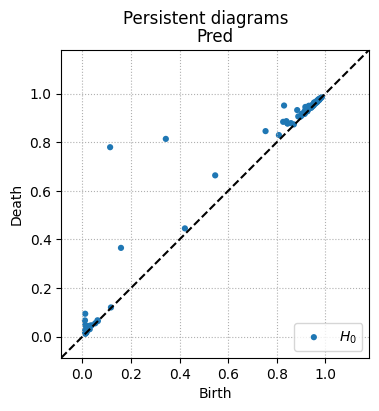}}%
  \caption{Persistence diagrams for 5 (left) and 95 (right) epochs}
  \label{f:2}
\end{figure}

We trained U-Net with and without topological priors on 100 epochs using Adam optimizer. Examples of changing persistence diagrams for prediction during training with topological loss are on Fig. \ref{f:2}. We can see that number of points decreased.

\subsection{Results}

\begin{figure}
    \centering
    \includegraphics[scale=0.5,width=0.97\linewidth]{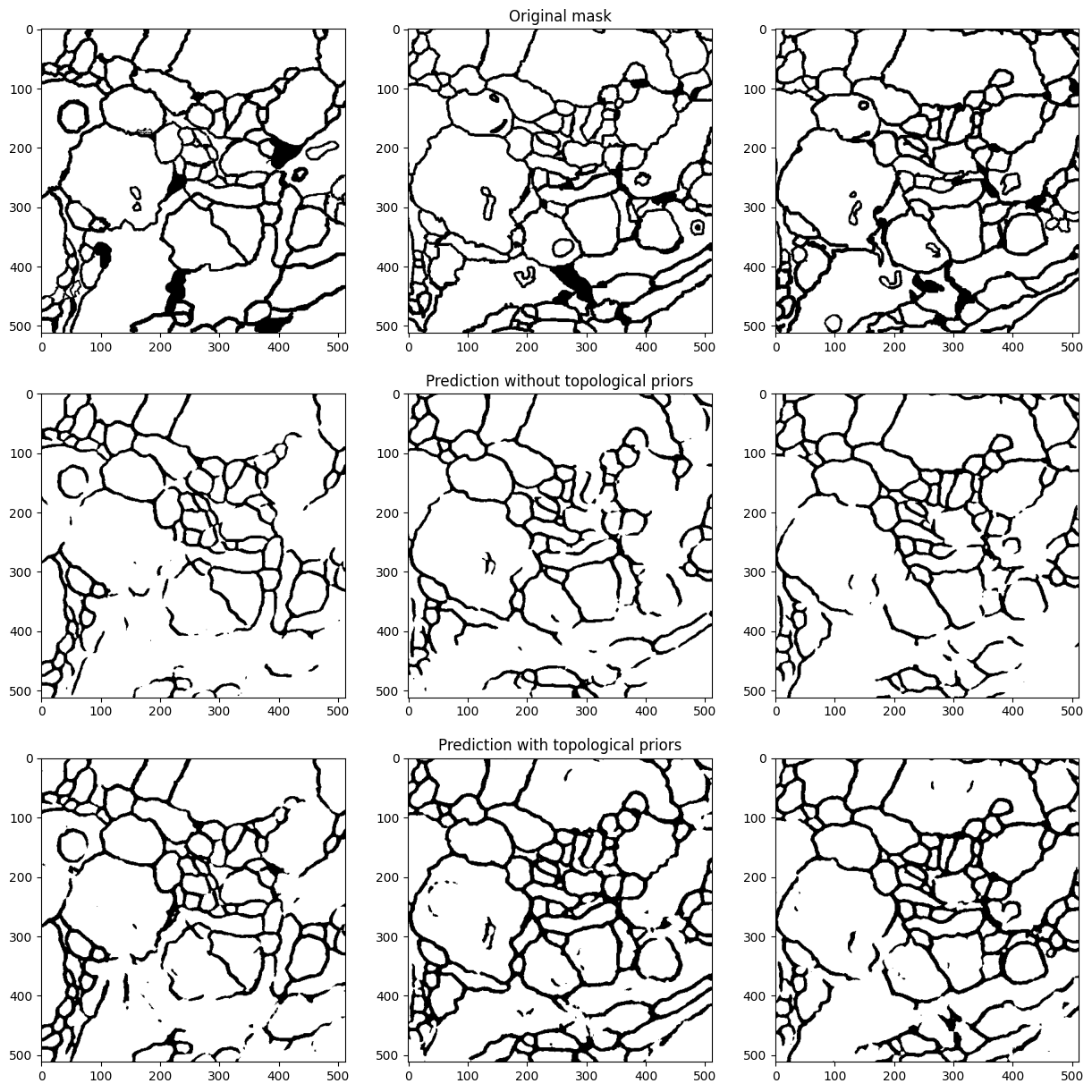}
  \caption{Predictions without topological priors (middle row) and with topoloss (bottom row)}
  \label{f:3}
\end{figure}

After training on small patches, we plotted predictions on whole images. The comparison of predictions of U-Net trained without any topological priors with trained with a topological loss on Fig. \ref{f:3}. 

\begin{figure}[H]
    \centering
    \includegraphics[scale=0.5,width=0.97\linewidth]{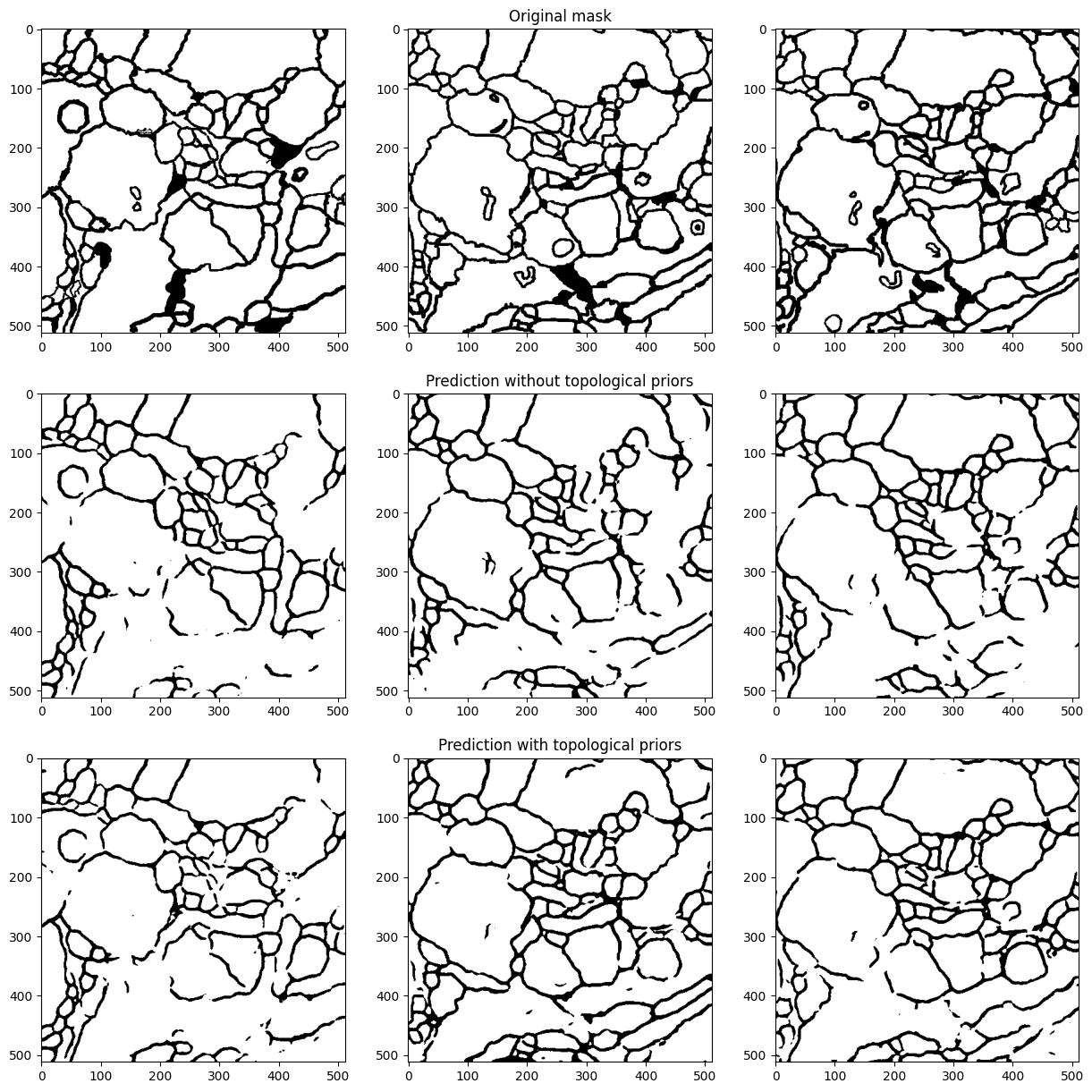}
  \caption{Predictions without topological priors (middle row) and with topological image processing (bottom row)}
  \label{f:4}
\end{figure}

Also, the comparison of predictions of U-Net trained without any topological priors with trained with a topological input image processing on Fig. \ref{f:4}. We can see that using topological priors make a prediction with a smaller amount of holes in cells.

Finally, we compute all metrics for various techniques and summarise the findings in table \ref{sample-table}. From metrics, it can be seen that topological priors give better results than classical segmentation (Simple UNet) in terms of accuracy, quality, and correctness, as well as topological metric (Betti-error).

\begin{table}[]
\resizebox{0.5\textwidth}{!}{%
\begin{tabular}{|
>{\columncolor[HTML]{EFEFEF}}c |c|c|c|}
\hline
\cellcolor[HTML]{EFEFEF} & \cellcolor[HTML]{EFEFEF} & \cellcolor[HTML]{EFEFEF} & \cellcolor[HTML]{EFEFEF} \\
\cellcolor[HTML]{EFEFEF} & \cellcolor[HTML]{EFEFEF} & \cellcolor[HTML]{EFEFEF} & \cellcolor[HTML]{EFEFEF} \\
\multirow{-3}{*}{\cellcolor[HTML]{EFEFEF}\textit{Metric}} & \multirow{-3}{*}{\cellcolor[HTML]{EFEFEF}\textit{Simple UNet}} & \multirow{-3}{*}{\cellcolor[HTML]{EFEFEF}\textit{\begin{tabular}[c]{@{}c@{}}UNet with \\ topological \\ priors in  training\end{tabular}}} & \multirow{-3}{*}{\cellcolor[HTML]{EFEFEF}\textit{\begin{tabular}[c]{@{}c@{}}UNet with  \\ topological \\ image processing\end{tabular}}} \\ \hline
\textit{Accuracy} & 0.899 & \textbf{0.910} & 0.909 \\ \hline
\textit{Completeness} & \textbf{0.975} & 0.951 & 0.971 \\ \hline
\textit{Correctness} & 0.907 & \textbf{0.938} & 0.92 \\ \hline
\textit{Quality} & 0.886 & 0.894 & \textbf{0.895} \\ \hline
\textit{Dice} & 0.939 & 0.944 & \textbf{0.945} \\ \hline
\textit{Betti error} & 1.16 & \textbf{1.02} & 1.1 \\ \hline
\end{tabular}%
}
\caption{Results for different strategies.}
\label{sample-table}
\end{table}

\section{Conclusion}

We have shown that incorporating the topological priors into deep neural networks (UNet) is a worthwhile idea, especially for problems whose solutions are primarily dependent upon the geometry of the underlying data. We have also shown the usefulness of posing topological information at different stages in the architecture for supervised learning problems. The results of the experiments themselves demonstrate the superiority of the concept that was presented and should be able to discover a variety of applications in data analysis and other prospective domains that are linked. Despite the fact that our findings were encouraging, the training process is somewhat slow, which limits its applicability to datasets that contain multiple variables. So, the plan for the future is to improve the computational efficiency of this method and apply it to multi-dimensional datasets like 3D segmentation tasks. This project's source code is also accessible \footnote{\url{https://github.com/ShakirSofi/TopoSeg.git}}


\bibliographystyle{IEEEtran}
\bibliography{bibtek}

\begin{thebibliography}{10}
\providecommand{\url}[1]{#1}
\csname url@samestyle\endcsname
\providecommand{\newblock}{\relax}
\providecommand{\bibinfo}[2]{#2}
\providecommand{\BIBentrySTDinterwordspacing}{\spaceskip=0pt\relax}
\providecommand{\BIBentryALTinterwordstretchfactor}{4}
\providecommand{\BIBentryALTinterwordspacing}{\spaceskip=\fontdimen2\font plus
\BIBentryALTinterwordstretchfactor\fontdimen3\font minus
  \fontdimen4\font\relax}
\providecommand{\BIBforeignlanguage}[2]{{%
\expandafter\ifx\csname l@#1\endcsname\relax
\typeout{** WARNING: IEEEtran.bst: No hyphenation pattern has been}%
\typeout{** loaded for the language `#1'. Using the pattern for}%
\typeout{** the default language instead.}%
\else
\language=\csname l@#1\endcsname
\fi
#2}}
\providecommand{\BIBdecl}{\relax}
\BIBdecl

\bibitem{mosin}
A.~Mosinska, P.~Marquez-Neila, M.~Kozinski, and P.~Fua, ``Beyond the pixel-wise
  loss for topology-aware delineation,'' 2017.

\bibitem{okty}
\BIBentryALTinterwordspacing
O.~Oktay, E.~Ferrante, K.~Kamnitsas, M.~Heinrich, W.~Bai, J.~Caballero, S.~A.
  Cook, A.~de~Marvao, T.~Dawes, D.~P. O’Regan, and et~al., ``Anatomically
  constrained neural networks (acnns): Application to cardiac image enhancement
  and segmentation,'' \emph{IEEE Transactions on Medical Imaging}, vol.~37,
  no.~2, p. 384–395, Feb 2018. [Online]. Available:
  \url{http://dx.doi.org/10.1109/TMI.2017.2743464}
\BIBentrySTDinterwordspacing

\bibitem{Chen}
\BIBentryALTinterwordspacing
C.~Chen, X.~Ni, Q.~Bai, and Y.~Wang, ``A topological regularizer for
  classifiers via persistent homology,'' in \emph{Proceedings of the
  Twenty-Second International Conference on Artificial Intelligence and
  Statistics}, ser. Proceedings of Machine Learning Research, K.~Chaudhuri and
  M.~Sugiyama, Eds., vol.~89.\hskip 1em plus 0.5em minus 0.4em\relax PMLR,
  16--18 Apr 2019, pp. 2573--2582. [Online]. Available:
  \url{https://proceedings.mlr.press/v89/chen19g.html}
\BIBentrySTDinterwordspacing

\bibitem{tiip}
\BIBentryALTinterwordspacing
N.~G. Robin~Vandaele and O.~Gevaert, ``Topological image modification for
  object detection and topological image processing of skin lesions,'' in
  \emph{Scientific Reports, nature 10:21061}, 2020. [Online]. Available:
  \url{https://doi.org/10.1038/S41598-020-77933-Y}
\BIBentrySTDinterwordspacing

\bibitem{assaf}
\BIBentryALTinterwordspacing
R.~Assaf, A.~Goupil, V.~Vrabie, and M.~Kacim, ``{Homology Functionality for
  Grayscale Image Segmentation},'' in \emph{{CRIMSTIC}}, ser. Current Research
  in Information Technology, Mathematical Sciences, vol.~8, Melaka, Malaysia,
  2016, pp. 281--286. [Online]. Available:
  \url{https://hal.archives-ouvertes.fr/hal-02108426}
\BIBentrySTDinterwordspacing

\bibitem{clough1}
J.~R. Clough, I.~Oksuz, N.~Byrne, J.~A. Schnabel, and A.~P. King, ``Explicit
  topological priors for deep-learning based image segmentation using
  persistent homology,'' 2019.

\bibitem{clough2}
J.~Clough, N.~Byrne, I.~Oksuz, V.~A. Zimmer, J.~A. Schnabel, and A.~King, ``A
  topological loss function for deep-learning based image segmentation using
  persistent homology,'' \emph{IEEE Transactions on Pattern Analysis and
  Machine Intelligence}, pp. 1--1, 2020.

\bibitem{ying}
Y.-J. Xin and Y.-H. Zhou, ``Topology on image processing,'' in
  \emph{Proceedings of ICSIPNN '94. International Conference on Speech, Image
  Processing and Neural Networks}, 1994, pp. 764--767 vol.2.

\bibitem{xiao1}
X.~Hu, L.~Fuxin, D.~Samaras, and C.~Chen, ``Topology-preserving deep image
  segmentation,'' 2019.

\bibitem{Edelph}
H.~Edelsbrunner and J.~Harer, ``Persistent homology — a survey.''

\bibitem{Otterroadmap}
\BIBentryALTinterwordspacing
N.~Otter, M.~A. Porter, U.~Tillmann, P.~Grindrod, and H.~A. Harrington, ``A
  roadmap for the computation of persistent homology,'' \emph{EPJ Data
  Science}, vol.~6, no.~1, Aug 2017. [Online]. Available:
  \url{http://dx.doi.org/10.1140/epjds/s13688-017-0109-5}
\BIBentrySTDinterwordspacing

\bibitem{assaf2}
R.~Assaf, A.~Goupil, V.~Vrabie, T.~Boudier, and M.~Kacim, ``Persistent homology
  for object segmentation in multidimensional grayscale images,'' \emph{Pattern
  Recognition Letters}, vol. 112, pp. 277--284, 09 2018.

\bibitem{Rieck}
B.~Rieck and H.~Leitte, ``Agreement analysis of quality measures for
  dimensionality reduction,'' in \emph{Topological Methods in Data Analysis and
  Visualization IV: Theory, Algorithms, and Applications}.\hskip 1em plus 0.5em
  minus 0.4em\relax Springer International Publishing, 2017, pp. 103--117.

\bibitem{Heip}
C.~Heipke, H.~Mayer, C.~Wiedemann, and O.~Jamet, ``Empirical evaluation of
  automatically extracted road axes,'' 1998.

\end{thebibliography}

\end{document}